\documentclass{IOS-Book-Article}

\usepackage{mathptmx}

\usepackage[utf8]{inputenc} 
\RequirePackage{hyperref}
\RequirePackage{hypernat}

\usepackage{graphicx}
\usepackage{url}
\usepackage{amssymb}
\usepackage[dvipsnames]{xcolor}
\usepackage{enumerate}
\usepackage{algpseudocode, algorithm}

\usepackage{caption}
\usepackage{subcaption}
\usepackage{multirow}
\usepackage{rotating}

\def\hb{\hbox to 10.7 cm{}}

\begin{document}

\pagestyle{headings}
\def\thepage{}

\begin{frontmatter}              % The preamble begins here.

%\pretitle{Pretitle}
\title{Knowledge Graphs on the Web --\\an Overview}

\markboth{}{January 2020\hb}
%\subtitle{Subtitle}

\author{\fnms{Nicolas} \snm{Heist}}, 
\author{\fnms{Sven} \snm{Hertling}}, 
\author{\fnms{Daniel} \snm{Ringler}}, 
and
\author{\fnms{Heiko} \snm{Paulheim}}

\runningauthor{Heist et al.}
\address{Data and Web Science Group, University of Mannheim, Germany}

\begin{abstract}
Knowledge Graphs are an emerging form of knowledge representation.
While Google coined the term \emph{Knowledge Graph} first and promoted it as a means to improve their search results, they are used in many applications today.
In a knowledge graph, entities in the real world and/or a business domain (e.g., people, places, or events) are represented as nodes, which are connected by edges representing the relations between those entities.
While companies such as Google, Microsoft, and Facebook have their own, non-public knowledge graphs, there is also a larger body of publicly available knowledge graphs, such as DBpedia or Wikidata.
In this chapter, we provide an overview and comparison of those publicly available knowledge graphs, and give insights into their contents, size, coverage, and overlap.
\end{abstract}

\end{frontmatter}
\markboth{January 2020\hb}{January 2020\hb}

\section{Introduction}
Knowledge Graphs are increasingly used as means to represent knowledge. Due to their versatile means of representation, they can be used to integrate different heterogeneous data sources, both within as well as across organizations. \cite{galkin2016enterprise,gomez2017enterprise}

Besides such domain-specific knowledge graphs which are typically developed for specific domains and/or use cases, there are also public, cross-domain knowledge graphs encoding common knowledge, such as DBpedia, Wikidata, or YAGO. \cite{ringler2017one}
Such knowledge graphs may be used, e.g., for automatically enriching data with background knowledge to be used in knowledge-intensive downstream applications. \cite{ristoski2016semantic}
In particular for the case of explainable AI, knowledge graphs can be used as additional input to the AI algorithm, as a means to support interpretation of the results, or both. \cite{lecue2019role}

Since Google coined the term \emph{Knowledge Graph} for marketing purposes, it has subsequently been used in the scientific literature as well. The slogan by which Google announced KGs was \emph{Things, not Strings}.\footnote{\url{https://www.blog.google/products/search/introducing-knowledge-graph-things-not/}} 
The idea of that slogan is: while strings are often ambiguous, knowledge graphs consist of disambiguated entities, so that entities of the same name can be told apart more easily.
Nowadays, almost all companies processing large amounts of heterogeneous data use knowledge graphs as a means of representation, including, but not limited to IBM, Microsoft, Facebook or Ebay. \cite{noy2019industry}.

There are quite a few different definitions for knowledge graphs. \cite{ehrlinger2016towards}. Typically, a knowledge graph 
\begin{enumerate}
	\item mainly describes real world entities and their interrelations, organized in a graph.
	\item defines possible classes and relations of entities in a schema.
	\item allows for potentially interrelating arbitrary entities with each other.
	\item covers various topical domains. \cite{paulheim2017kg}
\end{enumerate}

In this chapter, we provide an overview of publicly available, cross-domain knowledge graphs on the Web. We discuss the techniques used to create those knowledge graphs and provide an in-depth comparison in terms of size, level of detail, contents, and overlap.

\section{Overview}
There are different techniques for creating knowledge graphs. The most common ones are (1) manual curation, (2) creation from (semi) structured sources, and (3) creation from unstructured sources. Some knowledge graphs also use a mix of those techniques.

\subsection{Manual Curation}
\emph{Cyc} \cite{lenat1995cyc} is one of the oldest knowledge graphs; the Cyc project dates back to the 1990s. Cyc was created along with its own language (CycL), which provides a large degree of formalization.

While Cyc was developed by a comparatively small group of experts, the idea of \emph{Freebase} \cite{pellissier2016freebase} was to establish a large community of volunteers, compared to Wikipedia. To that end, the schema of Freebase was kept fairly simple to lower the entrance barrier as much as possible. Freebase was acquired by Google in 2010 and shut down in 2014.

\emph{Wikidata} \cite{vrandevcic2014wikidata} also uses on a crowd editing approach. In contrast to Cyc and Freebase, Wikidata also imports entire whole large datasets, such as several national libraries' bibliographies. Porting the data from Freebase to Wikidata is also a long standing goal \cite{pellissier2016freebase}.

Curating a knowledge graph manually can be a large effort. The total cost of development for Cyc have been estimated as 120 Million USD.\footnote{\url{http://www.ttivanguard.com/conference/Napa2017/4-Lenat.pdf}}
This corresponds to a total cost of 2-6 USD per single axiom in Cyc. \cite{paulheim2018much}.

\subsection{Creation from (Semi) Structured Sources}
A more efficient way of knowledge graph creation is the use of structured or semi structured sources. Wikipedia is a commonly used starting point for knowledge graphs such as \emph{DBpedia} \cite{LehmannDBpedia} and \emph{YAGO} \cite{SuchanekYAGO}. 

\emph{DBpedia} mainly uses infoboxes in Wikipedia. Those are manually mapped to a pre-defined ontology; the mapping is crowd sourced using a Wiki and a community of volunteers. Given those mappings, the DBpedia Extraction Framework creates a graph in which each page in Wikipedia becomes an entity, and all values and links in an infobox become attributes and edges in the graph.

\emph{YAGO} uses a similar process, but classifies instances based on the category structure and WordNet \cite{miller1995wordnet} instead of infoboxes. YAGO integrates various language editions of Wikipedia into a single graph and represents temporal facts with meta-level statements, i.e., RDF reification.

\emph{CaLiGraph} also uses information in categories, but aims at converting them into formal axioms using DBpedia as supervision \cite{heist2019uncovering}. Moreover, instances from Wikipedia list pages are considered for populating the knowledge graph \cite{paulheim2013extending}. The result is a knowledge graph which is not only richly populated on the instance level, but also has a large number of defining axioms for classes \cite{heist2020caligraph}.

A similar approach, i.e., the combination of information in Wikipedia and WordNet, is used by  \emph{BabelNet} \cite{navigli2012babelnet}. The main purpose of BabelNet is the collection of synonyms and translations in various languages, so that this knowledge graph is particularly well suited for supporting multi-language applications. Similarly, \emph{ConceptNet} \cite{speer2012representing} collects synonyms and translations in various languages, integrating multiple third party knowledge graphs itself.

\emph{DBkWik} \cite{hertling2018dbkwik} uses the same codebase as DBpedia, but applies it to a multitude of Wikis. This leads to a graph which has a larger coverage and level of detail for many long tail entities, and is highly complementary to DBpedia. However, the absence of a central ontology and mappings, as well as the existence of duplicates across Wikis, which might not be trivial to detect, imposes a number of challenges not present in DBpedia.

Another source of structured data is the structured annotations in Web pages using techniques such as RDFa, Microdata, and Microformats \cite{meusel2014webdatacommons}. While the pure collection of those could, in theory, already be considered a knowledge graph, that graph would be rather disconnected and consist of a plethora of small, unconnected components \cite{paulheim2015adoption} and would require additional cleanup for compensating irregular use of the underlying schemas and shortcomings in the extraction \cite{meusel2015heuristics}. A consolidated version of this data into a more connected knowledge graph has been published under the name \emph{VoldemortKG}  \cite{tonon2016voldemortkg}.

\subsection{Creation from Unstructured Sources}
The extraction of a knowledge graph from semi structured sources is considered more easy than from the extraction from unstructured sources. However, there is much more information in unstructured sources (such as text). Therefore, extracting knowledge from unstructured sources has also been proposed.

\emph{NELL} \cite{carlson2010coupled} is an example for extracting a knowledge graph from free text. NELL was originally trained with a few seed examples and continuously runs an iterative coupled learning process. In each iteration, facts are used to learn textual patterns to detect those facts, and patterns learned in previous iterations are used to extract new facts, which serve as training examples in later iterations. To improve the quality, NELL has introduced a feedback loop incorporating occasional human feedback.

\emph{WebIsA} \cite{seitner2016large} also extracts facts from free text, but focuses on the creation of a large-scale taxonomy. For each extracted fact, rich metadata are collected, including the sources, the original sentences, and the patterns used in the extraction of a particular fact. Those metadata are exploited for computing a confidence score for each fact. \cite{hertling2017webisalod}.

\section{Comparison of Knowledge Graphs}
Whenever a knowledge graph is to be used in an application, it is important to determine which knowledge graph is best suitable for an application at hand. The knowledge graphs mentioned above differ in their content, their level of detail, etc. Hence, in this chapter, we will discuss several characteristics of knowledge graphs and provide insights into the differences between them.

\subsection{General Metrics}
The most straightforward metrics to be used consider the mere amount of information contained in a knowledge graph. Measures that may be used include:
\begin{itemize}
    \item The number of instances in a graph
    \item The number of assertions (or edges between entities)
    \item The average and median linkage degree (i.e.: how many assertions per entity does the graph contain?)
\end{itemize}
As for using a knowledge graph in an XAI system, these metrics hint at the utility -- the more information about the domain at hand is present (i.e., the more instances are represented in the knowledge graph and the more detailed that information is), the more can an XAI application benefit in providing better results or better interpretations.

Another set of metrics can be defined for the schema or ontology level of a knowledge graph:
\begin{itemize}
    \item The number of classes defined in the schema
    \item The number of relations defined in the schema
    \item The average depth and width (branching factor) of the class hierarchy\footnote{While this could also be done for the property hierarchy, extensive property hierarchies are rather rare in common knowledge graphs.}
    \item The complexity of the schema
\end{itemize}
While the instance-based metrics focus more on the coverage of a domain in a knowledge graph, these schema-level metrics provide information about the richness and formality of that knowledge. They determine which techniques to use -- e.g., while more formal, very complex ontologies will call for using ontology reasoning, light-weight, but large-scale ontologies will be better exploited by statistical and distributional approaches.

Table~\ref{tab:statistics} depicts those metrics for some of the knowledge graphs discussed above. ConceptNet and WebIsA are not included, since they do not distinguish a schema and instance level (i.e., there is no specific distinction between a class and an instance), which does not allow for computing those metrics meaningfully. For Cyc, which is only available as a commercial product today, we used the free version OpenCyc, which has been available until 2017.\footnote{It is still available, e.g., at \url{https://github.com/asanchez75/opencyc}}

\begin{table}[t]
    \centering
    \caption{Basic Metrics of Open Knowledge Graphs}
    \label{tab:statistics}
    \begin{tabular}{|l|r|r|r|r|}
        \hline
         & DBpedia & YAGO & Wikidata & BabelNet \\
         \hline
         \# Instances & 5,044,223 & 6,349,359 & 52,252,549 & 7,735,436 \\
         \# Assertions & 854,294,312 & 479,392,870 & 732,420,508 & 178,982,397 \\
         Avg. linking degree & 21.30 & 48.26 & 6.38 & 0.00 \\
         Median ingoing edges & 0 & 0 & 0 & 0 \\
         Median outgoing edges & 30 & 95 & 10 & 9 \\
         \hline
         \# Classes & 760 & 819,292 & 2,356,259 & 6,044,564 \\
         \# Relations & 1355 & 77 & 6,236 & 22 \\
         Avg. depth of class tree & 3.51 & 6.61 & 6.43 & 4.11 \\
         Avg. branching factor of class tree & 4.53 & 8.48 & 36.48 & 71.0 \\
         Ontology complexity & SHOFD & SHOIF & SOD & SO \\
         \hline
        \hline
         & Cyc & NELL & CaLiGraph & Voldemort \\
         \hline
         \# Instances & 122,441 & 5,120,688 & 7,315,918 & 55,861 \\
         \# Assertions & 2,229,266 & 60,594,443 & 517,099,124 & 693,428 \\
         Avg. linking degree & 3.34 & 6.72 & 1.48 & 0 \\
         Median ingoing edges & 0 & 0 & 0 & 0 \\
         Median outgoing edges & 3 & 0 & 1 & 5 \\
         \hline
         \# Classes & 116,821 & 1,187 & 755,963 & 621 \\
         \# Relations & 148 & 440 & 271 & 294 \\
         Avg. depth of class tree & 5.58 & 3.13 & 4.74 & 3.17 \\
         Avg. branching factor of class tree & 5.62 & 6.37 & 4.81 & 5.40 \\
         Ontology complexity & SHOIFD & SROIF & SHOD & SH \\
         \hline    \end{tabular}
\end{table}

From those metrics, it can be observed that the KGs differ in size by several orders of magnitude. The sizes range from 50,000 instances (and Voldemort) to 50 million instances (for Wikidata), so the latter is larger by a factor of 1,000. The same holds for assertions. Concerning the linkage degree, YAGO is much richer linked than the other graphs.

\begin{figure}[t]
    \centering
    \includegraphics[width=\textwidth]{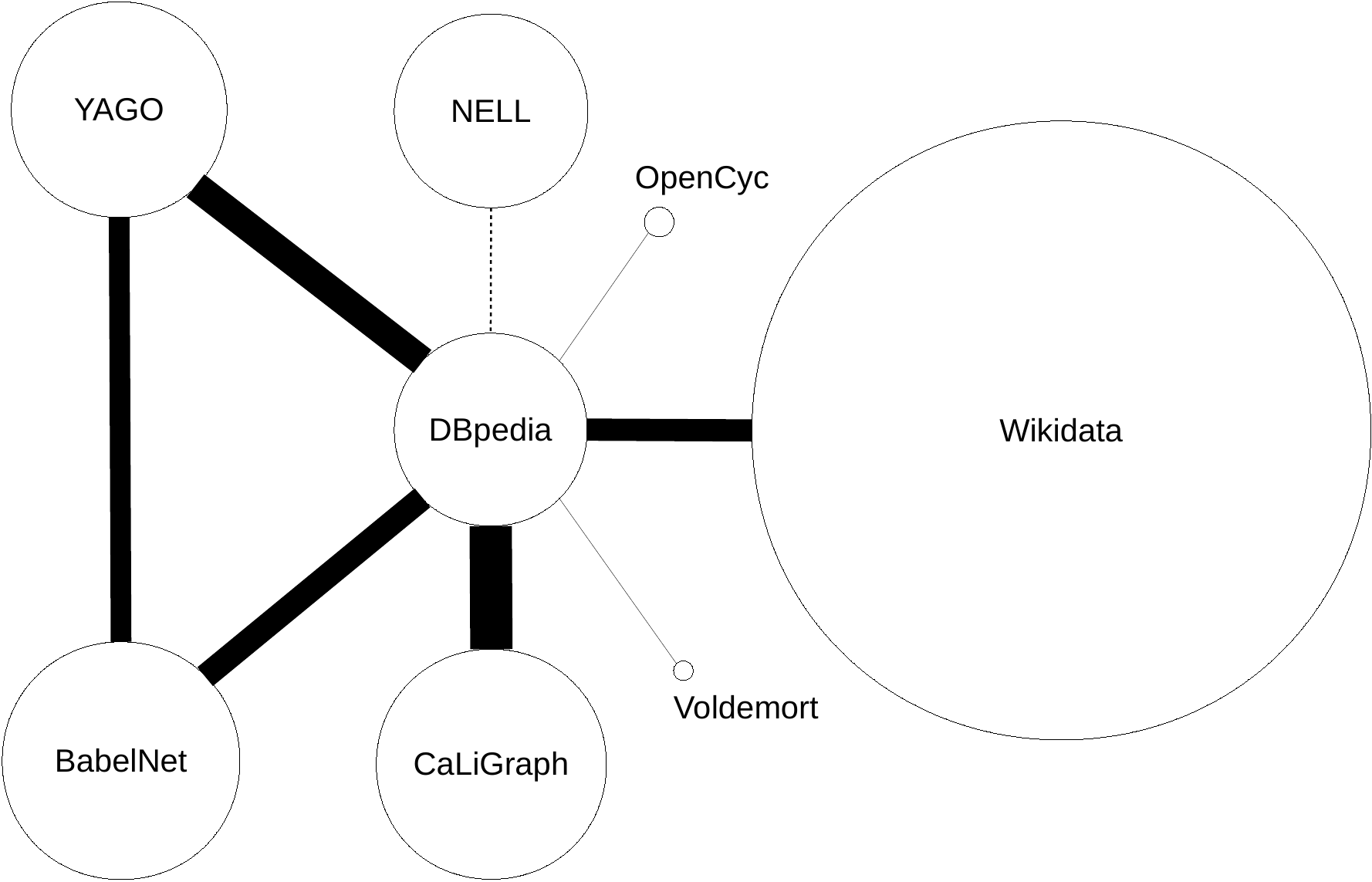}
    \caption{Depiction of the size and linkage degree of publicly available knowledge graphs. Although NELL and DBpedia are not explicitly interlinked, NELL contains links to Wikipedia, which can be trivially translated to DBpedia links.}
    \label{fig:graphs_overview}
\end{figure}

Figure~\ref{fig:graphs_overview} shows an overview of the knowledge graphs considered. 
We follow the conventions of the Linked Open Data Cloud diagrams\footnote{\url{https://www.lod-cloud.net/}} \cite{schmachtenberg2014adoption}, which are used to depict linked datasets and their connections.
In those diagrams, the size of the circles is proportional to the number of instances, and the strength of the connecting lines is proportional to the number of links.

The knowledge graphs also differ strongly in the characteristics of their schema. DBpedia and NELL have comparably small schemas, while Wikidata and BabelNet build deep and detailed taxonomies. For example, while NELL does not define detailed subclasses for \emph{Scientist}\footnote{\url{http://rtw.ml.cmu.edu/rtw/kbbrowser/pred:scientist}}, DBpedia defines four subclasses\footnote{\url{http://dbpedia.org/ontology/Scientist}}, Wikidata has more than 600\footnote{\url{https://www.wikidata.org/wiki/Q15976092}} and CaLiGraph almost 2,000\footnote{\url{http://caligraph.org/ontology/Scientist}}, including detailed classes such as \emph{sickle-cell disease researcher} or \emph{loop quantum gravity researcher}. Voldemort, on the other hand, reuses the schema.org ontology, which is comparably small~\cite{meusel2015web}.

Looking at the complexity, it is not much of a surprise that Cyc, originating in classic AI research and strongly building on logical rules \cite{buchanan2005very,lenat1995cyc}, has the highest complexity. Wikidata, BabelNet, and Voldemort have only little complexity, the other graphs are somewhere inbetween. 

\subsection{Contents}
\begin{figure}[t]
    \centering
	\includegraphics[width=\textwidth]{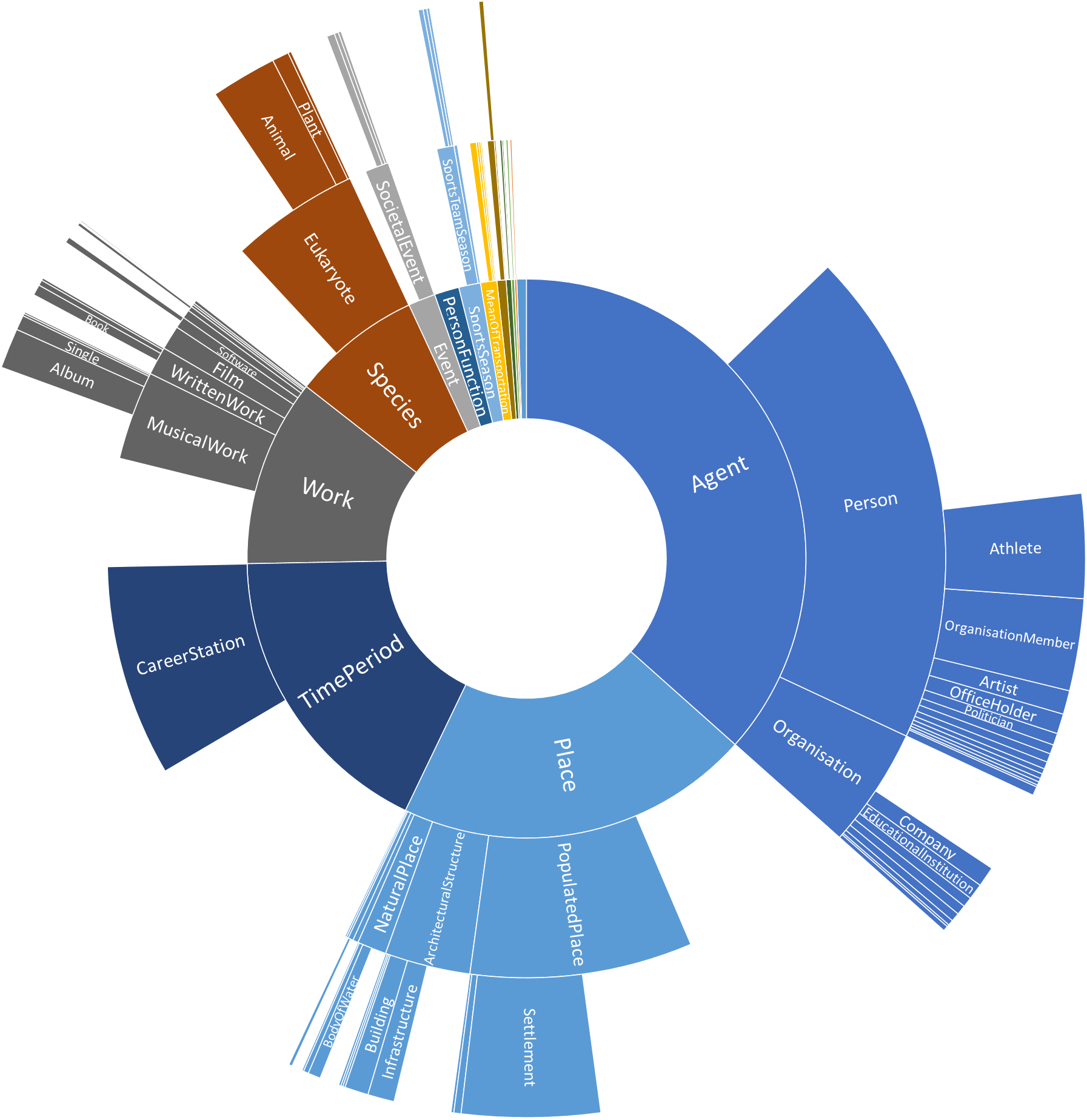}
	\caption{Instances in DBpedia}
	\label{fig:suburstDbpedia}
\end{figure}

\begin{figure}[t]
    \centering
	\includegraphics[width=\textwidth]{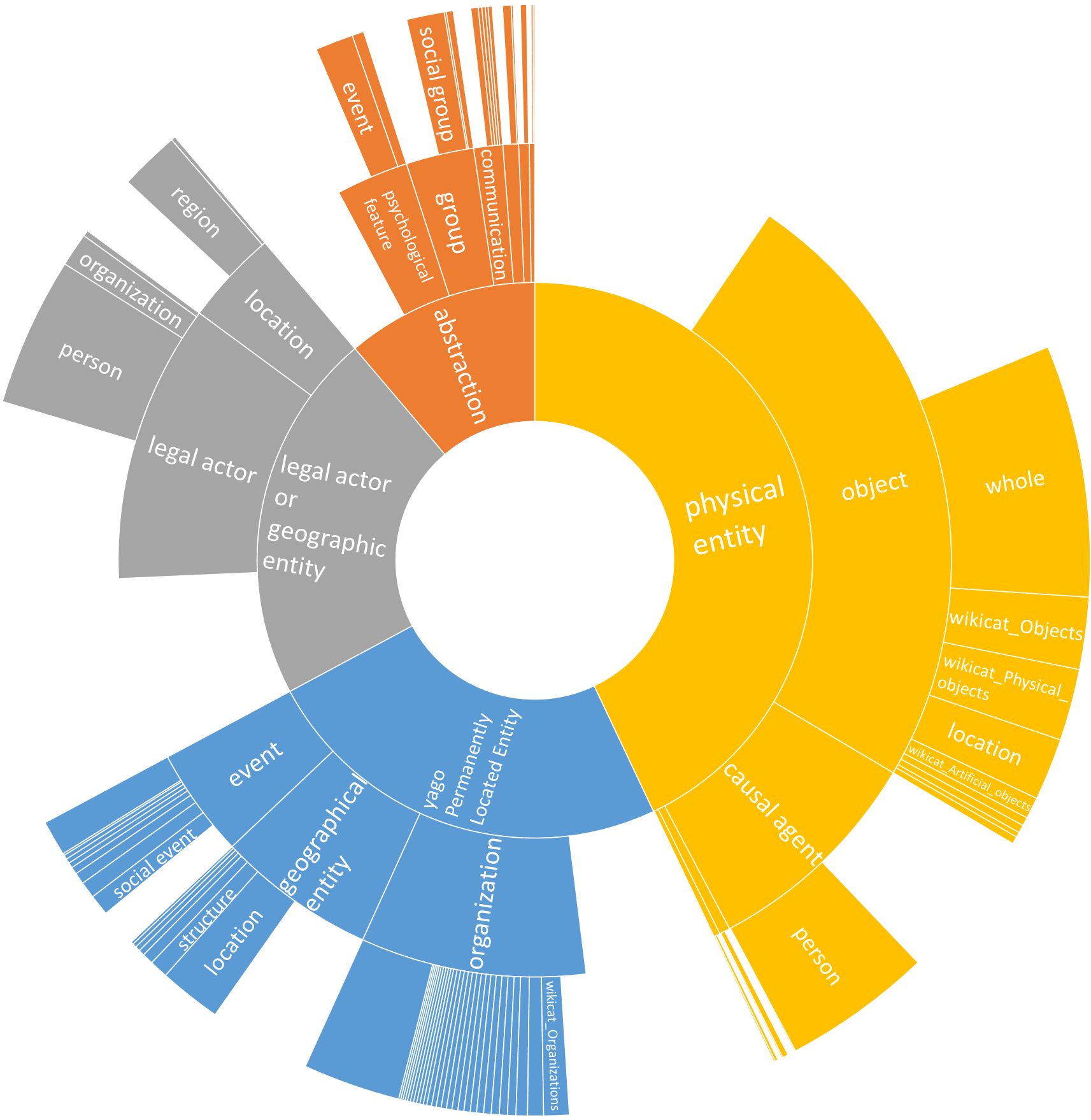}
	\caption{Instances in YAGO}
	\label{fig:suburstYAGO}
\end{figure}

\begin{figure}[t]
    \centering
	\includegraphics[width=\textwidth]{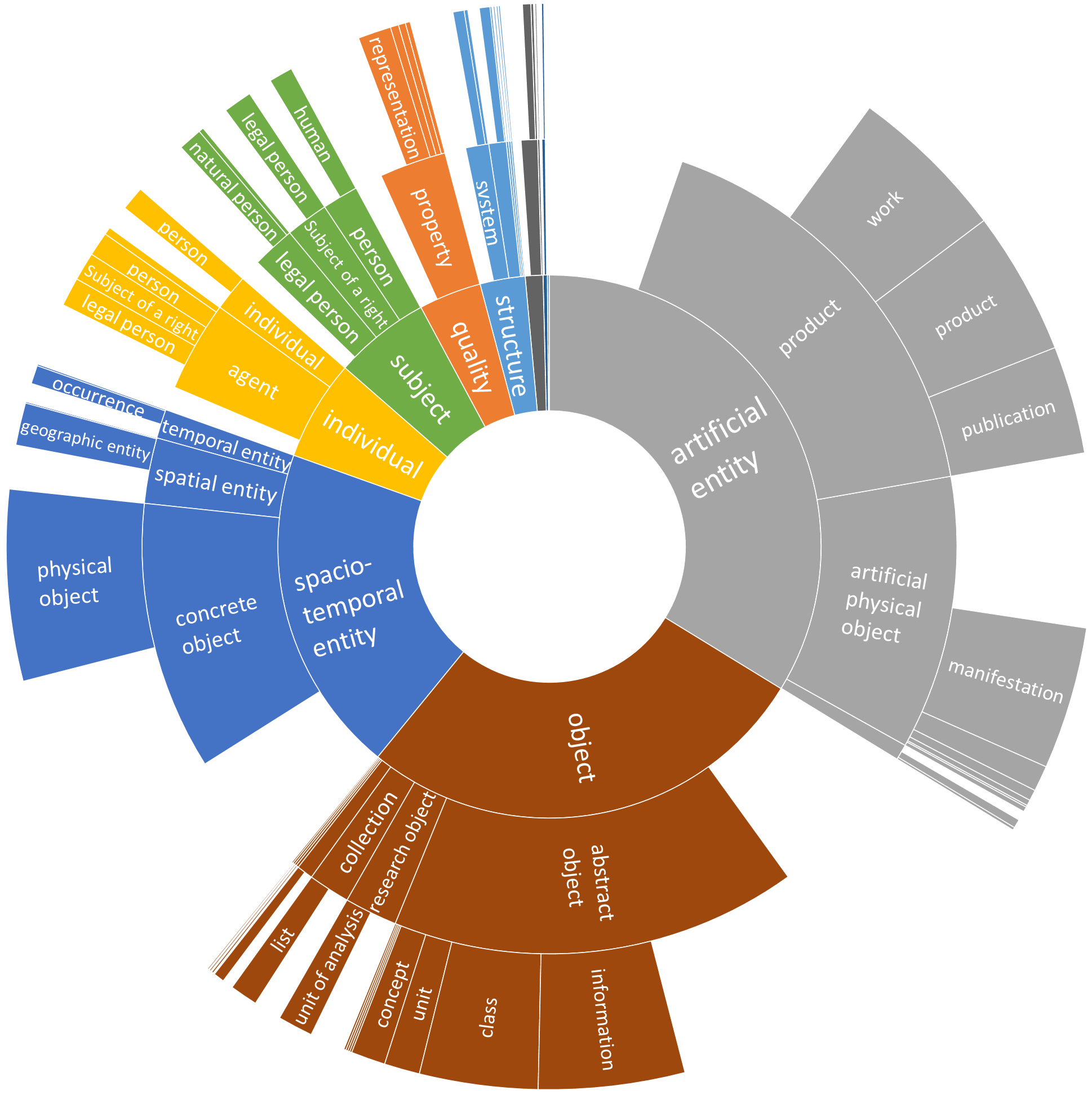}
	\caption{Instances in Wikidata}
	\label{fig:suburstWikidata}
\end{figure}

\begin{figure}[t]
    \centering
	\includegraphics[width=\textwidth]{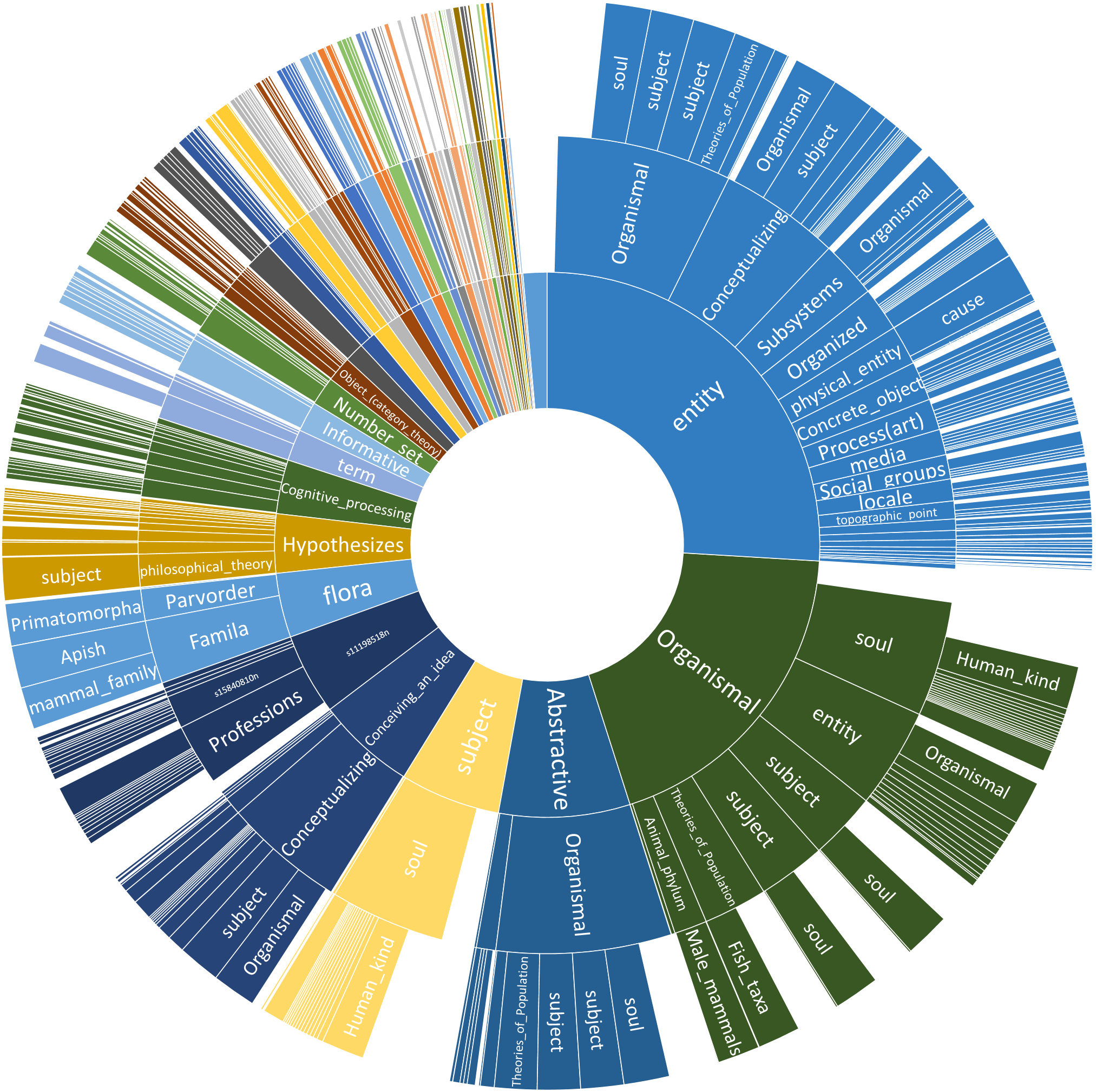}
	\caption{Instances in BabelNet}
	\label{fig:suburstBabelNet}
\end{figure}

\begin{figure}[t]
    \centering
	\includegraphics[width=\textwidth]{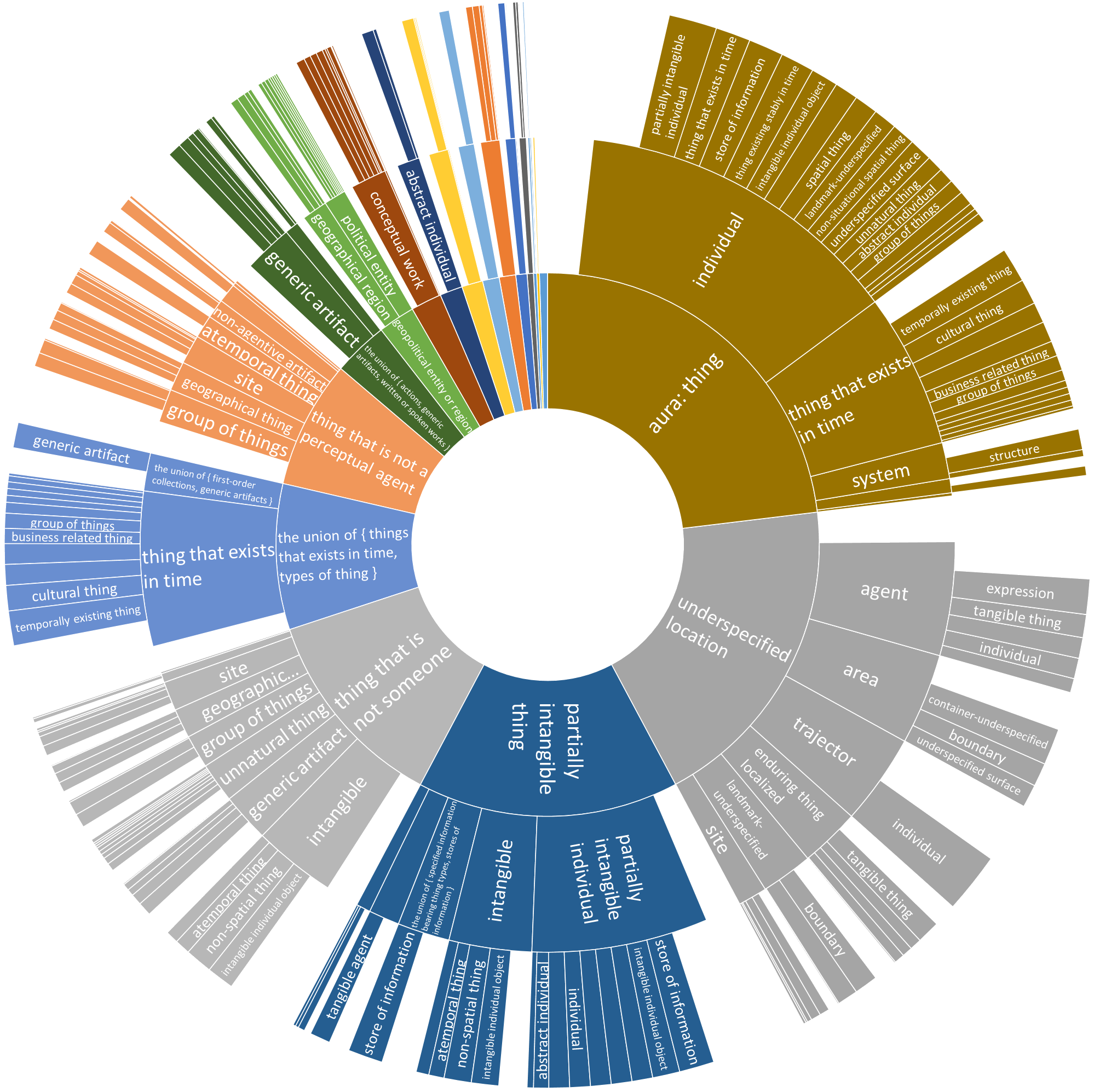}
	\caption{Instances in OpenCyc}
	\label{fig:suburstOpenCyc}
\end{figure}

\begin{figure}[t]
    \centering
	\includegraphics[width=\textwidth]{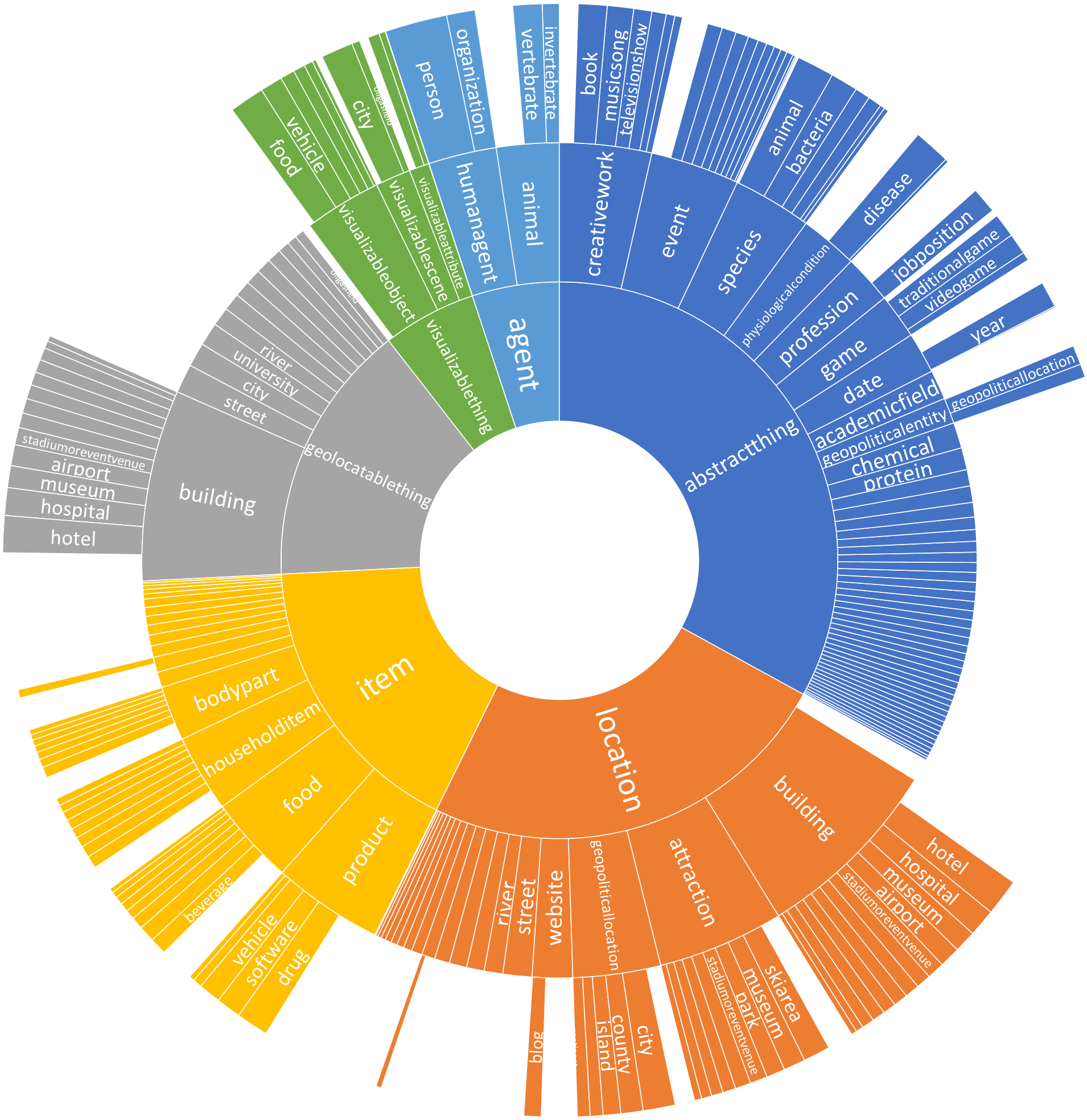}
	\caption{Instances in NELL}
	\label{fig:suburstNELL}
\end{figure}

\begin{figure}[t]
    \centering
	\includegraphics[width=\textwidth]{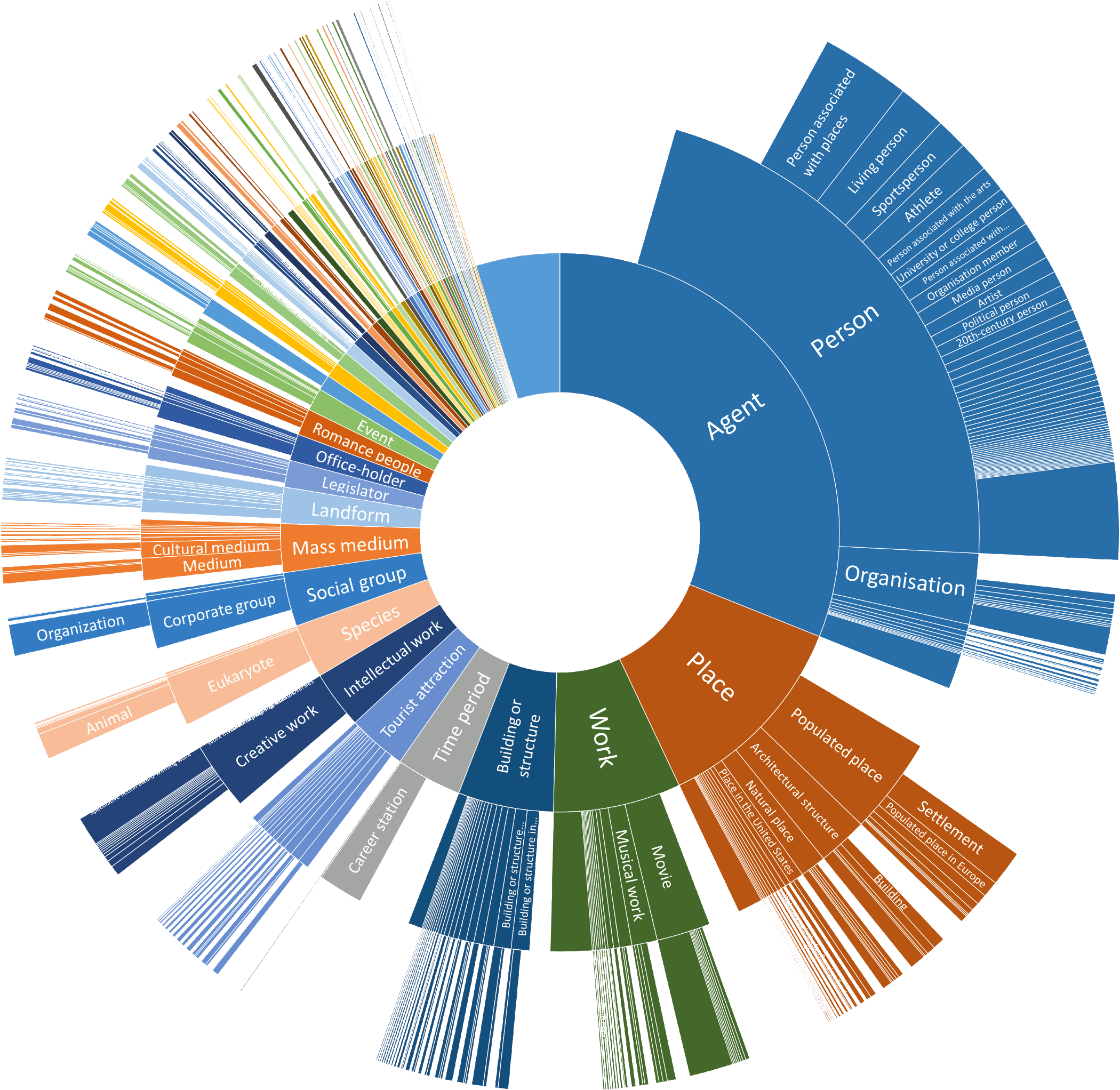}
	\caption{Instances in CaLiGraph}
	\label{fig:suburstCaLiGraph}
\end{figure}

\begin{figure}[t]
    \centering
	\includegraphics[width=0.7\textwidth]{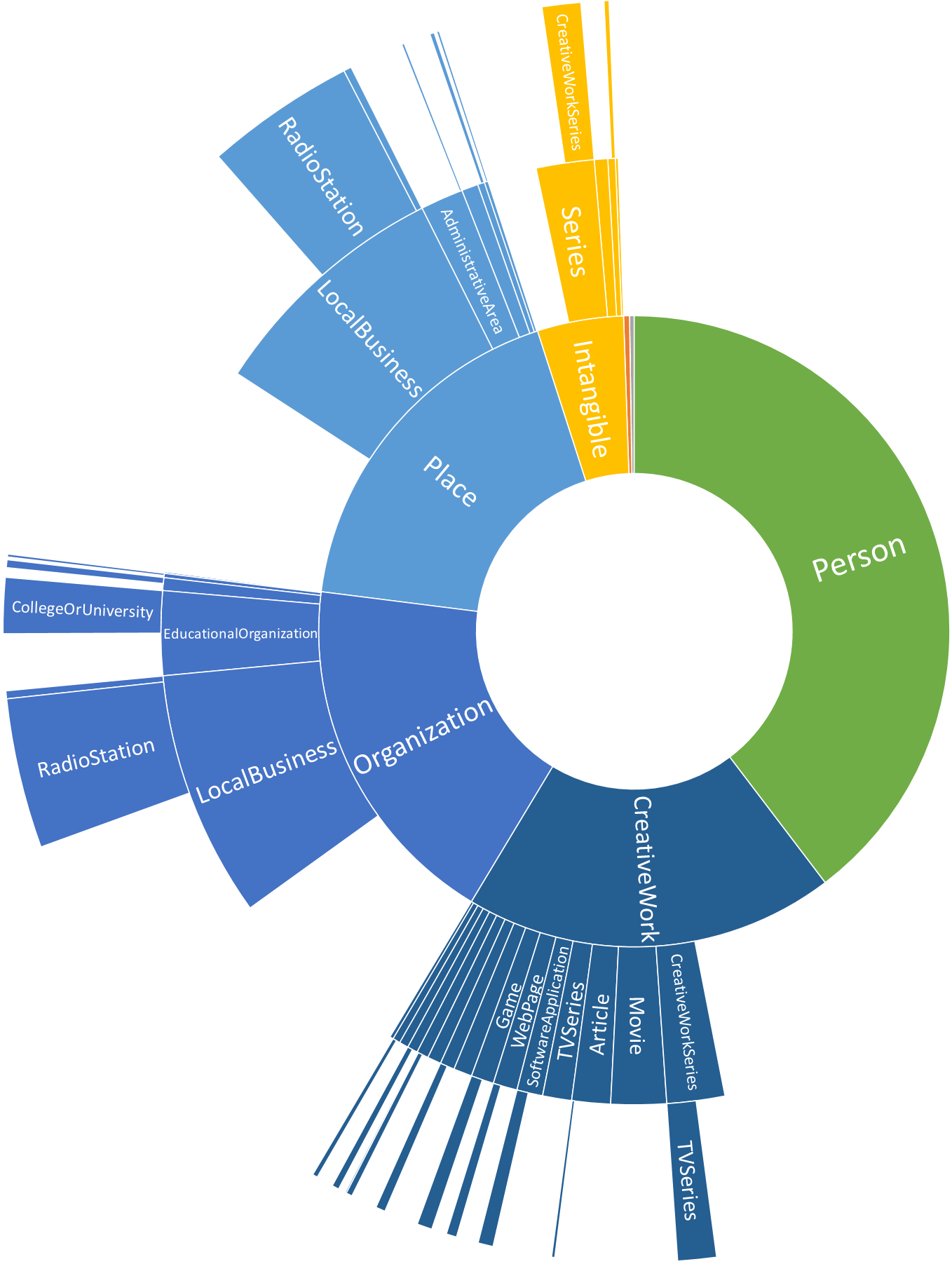}
	\caption{Instances in Voldemort}
	\label{fig:suburstVoldemort}
\end{figure}

The knowledge graphs do not only differ in their size and level of detail, but also in their contents. The most straightforward way to assess the content focus of a knowledge graph is to look at the size of its classes. Figures~\ref{fig:suburstDbpedia}-\ref{fig:suburstVoldemort} show graphic depictions of those class sizes. The diagrams were created starting from the most abstract class and following the class hierarchy to the largest respective subclasses.

At first glance, the figures reveal differences in the development of the taxonomies. While Cyc builds a formal ontology with very abstract top level categories such as \emph{partially intangible thing} or \emph{thing that exists in time}, the more pragmatic classification in DBpedia and Voldemort (the latter using schema.org as an ontology) has top level classes such as \emph{Place} or \emph{Person}. The reason for these differences lies in the origins of the respective knowledge graphs: While Cyc's classification was created by AI researchers, the ontology in DBpedia is the result of a crowdsourcing process \cite{paulheim2018much}. The same holds for schema.org, which is a pragmatic effort of a consortium of search engine developers.

Moreover, the diagrams reveal some differences in the contents. The main focus of DBpedia is on persons (and their careers), as well as places, works, and species. Wikidata also has a strong focus on works (mainly due to the import of entire bibliographic datasets), while Cyc, BabelNet and NELL show a more diverse distribution.

\subsection{Looking into Details}
To obtain deeper insights which classes are more prominent in which KGs, and, ultimately, which KGs are suitable for building explainable AI system in a specific domain, it is useful to not only look at the number of instances, but also the level of detail in which those instances are represented (i.e., the linkage degree and number of assertions per instance).

Table~\ref{tbl:detailed_view} depicts such a detailed view for ten prominent classes:
\begin{itemize}
    \item Person
    \item Organization
    \item Populated place (city, country, etc.)
    \item Uninhabited place  (mountain, lake, etc.)
    \item Species
    \item Work (book, movie, etc.)
    \item Building
    \item Gene
    \item Protein
    \item Event
\end{itemize}

\begin{sidewaystable}
\centering
\caption{Detail statistics for selected classes}
\label{tbl:detailed_view}
\footnotesize
\begin{tabular}{|l||r|r|r|r||r|r|r|r||r|r|r|r|}
    \hline
     & \multicolumn{4}{c||}{DBpedia} & \multicolumn{4}{c||}{YAGO} & \multicolumn{4}{c|}{Wikidata} \\
    Class & Instances & Avg. Deg. & Med-in & Med-out & Instances & Avg. Deg. & Med-in & Med-out & Instances & Avg. Deg. & Med-in & Med-out \\ 
    \hline
    Person & 1,243,400 & 1.54 & 0 & 5  & 2,213,431 & 5.62 & 0 & 258  & 5,250,840 & 2.41 & 0 & 10 \\
    Organization & 286,482 & 10.3 & 0 & 7  & 498,750 & 136.92 & 0 & 64  & 1,665,319 & 30.98 & 0 & 6 \\
    Populated place & 513,642 & 7.38 & 0 & 8  & 319,210 & 219.49 & 0 & 138  & 2,355,559 & 3.81 & 1 & 6 \\
    Uninhabited place & 67,495 & 0.91 & 0 & 4  & 160,615 & 23.67 & 0 & 48  & 1,516,890 & 0.23 & 0 & 6 \\
    Species & 306,104 & 2.56 & 0 & 7  & 2,553,369 & 4.92 & 0 & 224  & 110 & 21.53 & 0 & 5 \\
    Work & 496,070 & 0.81 & 0 & 8  & 1,175,125 & 28.07 & 0 & 36  & 34,585,828 & 5.91 & 0 & 12 \\
    Building & 197,831 & 0.41 & 0 & 5  & 274,606 & 13.44 & 0 & 69  & 2,291,168 & 0.98 & 0 & 6 \\
    Gene & 4 & 0.5 & 0.5 & 7.5  & 12,351 & 0.00 & 0 & 8  & 172,128 & 1.27 & 0 & 8 \\
    Protein & 2,747 & 0.03 & 0 & 1  & 10,935 & 0.01 & 0 & 52  & 84,163 & 1.15 & 1 & 14 \\
    Event & 76,029 & 1.91 & 0 & 3  & 562,583 & 41.23 & 0 & 48  & 579,559 & 1.76 & 0 & 5 \\
    \hline
    \hline
     & \multicolumn{4}{c||}{BabelNet} & \multicolumn{4}{c||}{Cyc} & \multicolumn{4}{c|}{NELL} \\
    Class & Instances & Avg. Deg. & Med-in & Med-out & Instances & Avg. Deg. & Med-in & Med-out & Instances & Avg. Deg. & Med-in & Med-out \\ 
    \hline
    Person & 2,384,065 & 0.00 & 0 & 17  & 12,784 & 0.04 & 0 & 3  & 90,601 & 8.93 & 0 & 0 \\
    Organization & 764,662 & 0.01 & 0 & 12  & 26,276 & 5.70 & 0 & 5  & 41,646 & 6.31 & 0 & 0 \\
    Populated place & 509,257 & 0.01 & 0 & 9  & 8,596 & 20.63 & 0 & 12  & 28,359 & 39.98 & 0 & 0 \\
    Uninhabited place & 70,209 & 0.02 & 0 & 11  & 64 & 2.05 & 1 & 12  & 158,879 & 3.83 & 0 & 0 \\
    Species & 6,536 & 0.01 & 0 & 17  & 0 & - & - & -  & 3,273 & 0.88 & 0 & 0 \\
    Work & 491,057 & 0.00 & 0 & 12  & 19,908 & 0.91 & 0 & 2  & 27,038 & 1.09 & 0 & 0 \\
    Building & 520 & 0.00 & 0 & 8  & 786 & 0.14 & 0 & 4  & 50,699 & 4.51 & 0 & 0 \\
    Gene & 522 & 0.00 & 0 & 5  & 8 & 0 & 0 & 3  & 0 & - & - & - \\
    Protein & 10,399 & 0.00 & 0 & 3  & 0 & - & - & -  & 0 & - & - & - \\
    Event & 9,904 & 0.00 & 0 & 13  & 685 & 0.86 & 0 & 2  & 37,203 & 0.65 & 0 & 0 \\
    \hline
    \hline
     & \multicolumn{4}{c||}{CaLiGraph} & \multicolumn{4}{c||}{Voldemort} \\
    Class & Instances & Avg. Deg. & Med-in & Med-out & Instances & Avg. Deg. & Med-in & Med-out \\ 
    \cline{1-9}
    Person & 1,967,339 & 0.34 & 0 & 2  & 36,370 & 0.00 & 0 & 5  \\
    Organization & 547,728 & 2.67 & 0 & 2  & 5,984 & 0.00 & 0 & 1 \\
    Populated place & 700,559 & 10.12 & 0 & 2  & 1,278 & 0.00 & 0 & 5 \\
    Uninhabited place & 170,324 & 1.16 & 0 & 2  & 60 & 0.00 & 0 & 4 \\
    Species & 552,249 & 1.04 & 0 & 1  & 0 & - & - & - \\
    Work & 678,888 & 0.49 & 0 & 1  & 6,673 & 0.00 & 0 & 3 \\
    Building & 404,087 & 0.21 & 0 & 1  & 108 & 0.00 & 0 & 5 \\
    Gene & 1,106 & 0.00 & 0 & 0  & 0 & - & - & - \\
    Protein & 6,138 & 0.00 & 0 & 0  & 0 & - & - & - \\
    Event & 148,122 & 0.49 & 0 & 0  & 198 & 0.00 & 0 & 4 \\
    \cline{1-9}
\end{tabular}
\end{sidewaystable}

The global trend observed in this table is that Wikidata has the largest number of instances in most of the classes, while YAGO has the largest level of detail. However, there are differences from class to class. While Wikidata has a large number of works, YAGO is a good source of events. NELL often has fewer instances, but a larger level of detail, which can be explained by its focus on more prominent instances. 

The contrast of the average and the median degree also reveals a few differences. For example, BabelNet contains a similar amount of instances as DBpedia for some classes, e.g., uninhabited places or works. While the average linkage degree is higher in DBpedia, the median is higher in BabelNet. This hints at a more uneven distribution of information in DBpedia, while BabelNet has a more constant distribution of statements per instance.

\section{Linkage and Overlap of Knowledge Graphs}
Since knowledge graphs differ so strongly in size, coverage, and level of detail, combining information from multiple KGs for implementing one application is often beneficial. To estimate the value of such a combination, we determine the overlap of the knowledge graphs first.

As shown in Fig.~\ref{fig:graphs_overview}, many KGs contain explicit interlinks. Those links, usually in the form of \texttt{owl:sameAs} links, express that entities in two KGs are the same (or, more precisely: that they refer to the same real world entity) \cite{halpin2010owl}. In other cases, such links can be generated indirectly, e.g., if a knowledge graph contains links to Wikipedia pages, which can be easily mapped to entities in DBpedia and YAGO.

Even if those links provide a first hint at the overlap of KGs, and further links can be found by exploiting the transitivity of the \texttt{owl:sameAs} property \cite{beek2018sameas}, they do not provide a complete picture. Due to the \emph{open world assumption}, which holds for KG interlinks as well, there might always be more links than the one which are explicitly or implicitly provided by the KGs.

\subsection{Method}
In order to estimate the actual number of interlinks, we use a method first discussed in \cite{ringler2017one}, which builds on a set of existing links and heuristic link discovery:
\begin{enumerate}
    \item We use different heuristics to discover links between two KGs automatically, e.g., different string similarity measures \cite{ferrara2011data,nentwig2017survey}.
    \item Based on the existing, incomplete set of interlinks, we measure recall and precision of the individual heuristics \cite{ritze2011towards}.
    \item With the help of those recall and precision figures, we can estimate the actual number of interlinks. After repeating the procedure with multiple heuristics, we can use the average of those estimations.
\end{enumerate}
Given that the actual number of links is $C$ (which is unknown), the number of links found by a heuristic is $F$, and that the number of correct links in $F$ is $F^+$, recall and precision are defined as
\begin{eqnarray}
R:= & \frac{\left|F^+\right|}{\left|C\right|} \label{eq:recall}\\
P:= & \frac{\left|F^+\right|}{\left|F\right|} \label{eq:precision}
\end{eqnarray}
By resolving both to $\left|F^+\right|$ and combining the equations, we can estimate $\left|C\right|$ as
\begin{equation}
\left|C\right| = \left|F\right| \cdot P \cdot \frac{1}{R}
\label{eq:estimate}
\end{equation}
Thus, we can obtain an estimate for $C$ given $F$, $R$, and $P$. A more intuitive interpretation of the last equation is that $P$ is a measure of how strongly the heuristic \emph{over}estimates the number of actual interlinks (thus, $F$ is reduced by multiplication with $P$), and $R$ is a measure of how strongly the heuristic \emph{under}estimates the number of actual interlinks (thus, $F$ is divided by $R$).

In \cite{ringler2017one}, we have shown that across different heuristics, although $F$ varies a lot, the estimate $C$ is fairly stable. For producing the estimates in this chapter, we have used the following heuristics: string equality, scaled Levenshtein (thresholds 0.8, 0.9, and 1.0), Jaccard (0.6, 0.8, and 1.0), Jaro (0.9, 0.95, and 1.0), JaroWinkler (0.9, 0.95, and 1.0), and MongeElkan (0.9, 0.95, and 1.0). The estimated overlap reported is the average estimate computed using these 16 metrics.

\subsection{Findings}
\begin{figure}[p]
	\centering
	\includegraphics[width=\linewidth]{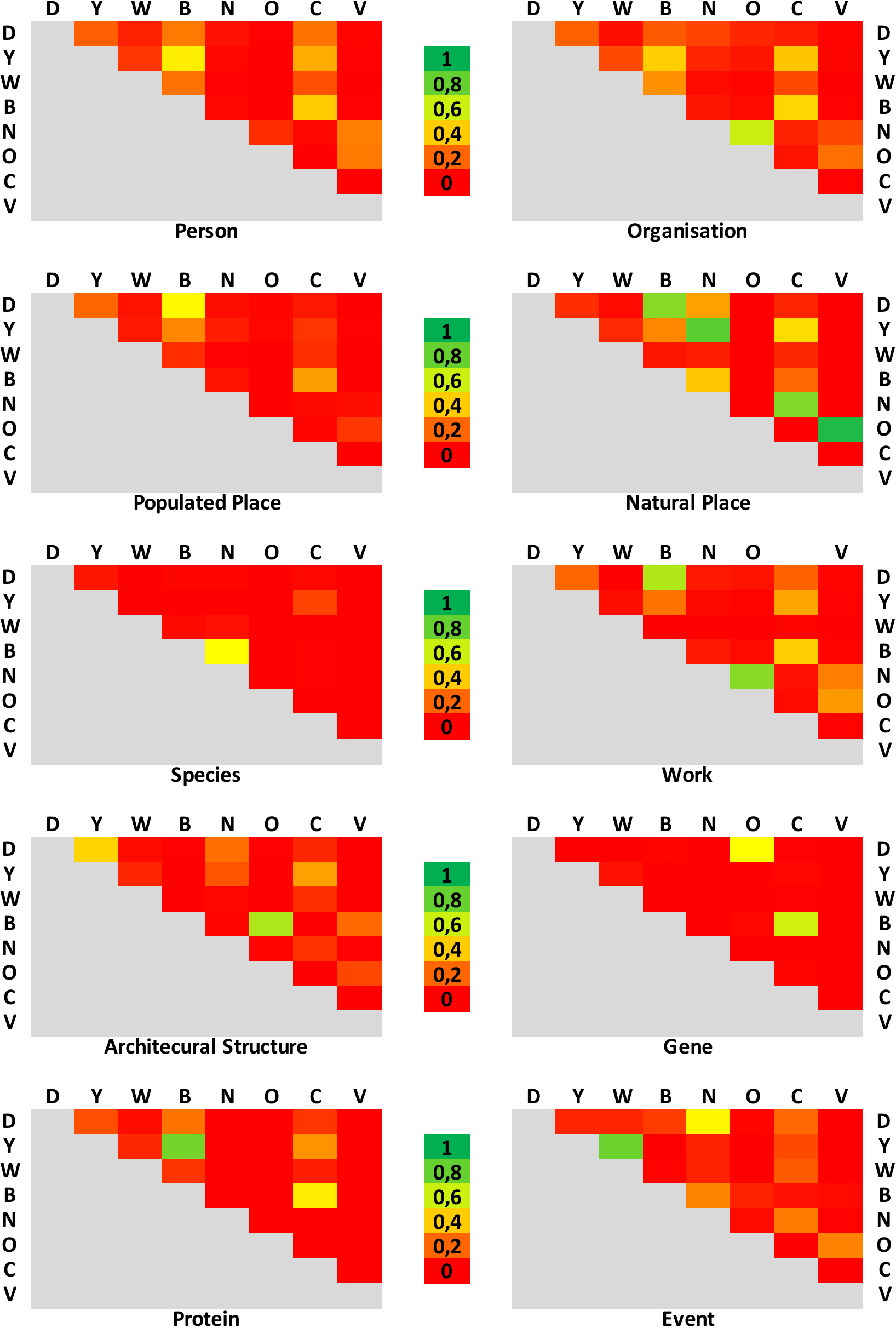}
	\caption{Fraction of entities in a pair of knowledge graphs which is \emph{not} contained in the larger of the two graphs.}
\label{fig:heatmaps-potential-gain}
\end{figure}

\begin{figure}[p]
	\centering
	\includegraphics[width=\linewidth]{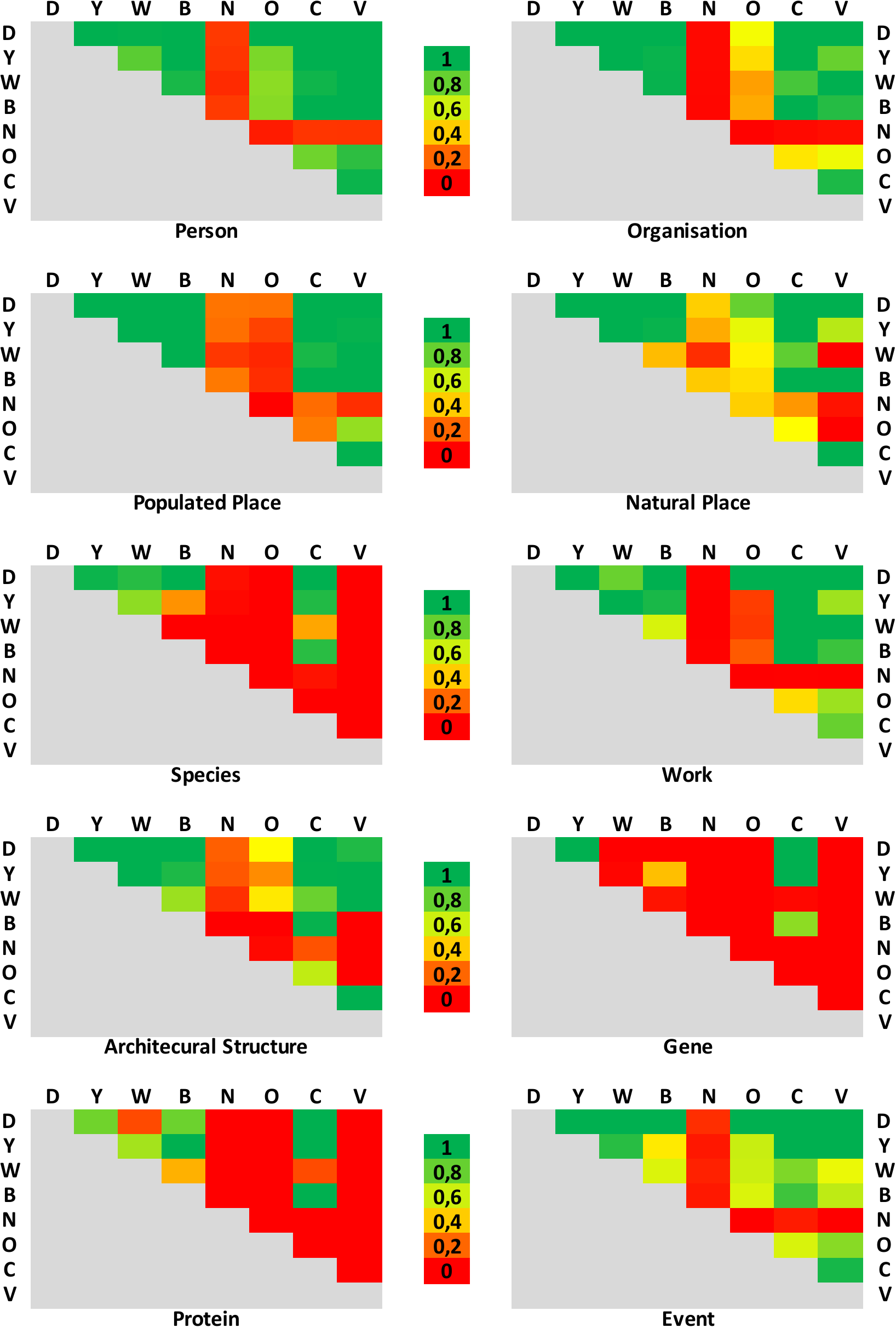}
	\caption{Existing entities in two KGs in relation to the number of links.}
	\label{fig:heatmaps-current-linkage}
\end{figure}

To analyze the benefit of the combination of different KGs, we depict the number of estimated links both in relation to (a) the entities existing in the larger of the two KGs (Fig.~\ref{fig:heatmaps-potential-gain}) as well as (b) in relation to the links that exist explicitly or implicitly (Fig.~\ref{fig:heatmaps-current-linkage}). From (a), we can estimate the amount of gain in knowledge of combining two KGs (i.e., if only a small fraction of one KG is also contained in the other and vice versa, such a combination adds a lot of information). From (b), we can get insights into whether the set of existing links is sufficient for such a combination or not.

Fig.~\ref{fig:heatmaps-potential-gain} shows that in most cases, the larger of two knowledge graphs contains most of the entities of a smaller one, i.e., its set of entities of a class in larger KG is usually a superset of that set in the smaller one. For example, as depicted in table~\ref{tbl:detailed_view}, Wikidata contains about twice as many persons as DBpedia and YAGO. A value close to 0 for the overlap implies that DBpedia and YAGO contain almost no persons which are not contained in Wikidata. In conclusion, combining Wikidata with DBpedia or YAGO for a better coverage of the \emph{Person} class would not be beneficial.

Notable exceptions are BabelNet and CaLiGraph, which often contain complementary instances. For example, DBpedia, BabelNet and CaLiGraph contain 1.2M, 2.4M, and 1.9M instances of the class \emph{Person}, respectively, while DBpedia and BabelNet together are estimated to 2.9M, and all three together are estimated to contain even 3.9M instances of the class \emph{Person}. The reasons for the high complementary of DBpedia/YAGO, BabelNet and CaLiGraph are their sources (only English Wikipedia vs. multiple language editions) and extraction mechanisms (especially the extraction from list pages in CaLiGraph, which leads to a larger number of instances overall).

Fig.~\ref{fig:heatmaps-current-linkage} shows that the linkage between DBpedia, YAGO, BabelNet and CaLiGraph is mostly complete (i.e., most of the common instances are also explicitly linked). Since they are all generated from Wikipedia with different means, this is not much surprising. On the other hand, Nell, OpenCyc, and Voldemort have a much lower degree of linkage. This shows that links between KGs are only complete where they are trivial to create, and combining different knowledge graphs otherwise requires efforts in improving the interlinking as a preliminary step.

\section{Conclusion and Outlook}
In this chapter, we have given an overview of publicly available, cross-domain knowledge graphs on the Web. We have compared them according to different metrics which might be helpful to implement an explainable AI project in a given domain.

Besides the metrics used for this comparison, there are quite a few more which help in the selection and assessment of a given KG. For example, data quality in KGs has not been considered in this chapter, since there are already quite elaborate surveys covering this aspect~\cite{farber2016linked,zaveri2016quality}.

So far, we have measured the overlap of knowledge graphs only based on entities. Another helpful metric would be the overlap on the statement level. Even if two knowledge graphs cover the same entity, the information they contain about that entity might still be complementary. For example, for the entity \emph{University of Mannheim}, DBpedia has the exact number of undergraduate students, PhD students, etc.\footnote{\url{http://dbpedia.org/page/University_of_Mannheim}}, while Wikidata lists all faculties\footnote{\url{https://www.wikidata.org/wiki/Q317070}} and can provide a list of researchers employed at the university\footnote{\url{https://w.wiki/7UU}}.
The density of information differs as well: while YAGO lists 3 alumni of the University of Mannheim\footnote{\url{https://bit.ly/2U4wL0A}}, DBpedia lists 11 and Wikidata even 85 alumni\footnote{\url{https://w.wiki/7UV}}. Even contradicting information can be found \cite{bryl2014interlinking}: for example, DBpedia and Wikidata provide a different number of students and Wikidata and YAGO provide different founding dates of the University of Mannheim.

Developing cross-domain knowledge graphs is an active field of research, and new developments emerge every once in a while. They differ in the data they use and/or the method of extraction:
\begin{itemize}
    \item \emph{DBkWik} \cite{hertling2018dbkwik,hertling2019dbkwik,hofmann2017dbkwik} uses the extraction mechanism of DBpedia and applies it to a multitude of Wikis. The intermediate result is a collection of a few thousand isolated knowledge graphs, which have to be integrated into a coherent joint knowledge graph.
    \item \emph{Chaudron} \cite{subercaze2017chaudron} uses Wikipedia as a source and focuses on quantifiable values (e.g., sizes, weights, etc.). Besides the mere extraction, Chaudron uses sophisticated methods for recognizing and converting units of measurement.
    \item The Linked Hypernym Dataset (\emph{LHD}) \cite{kliegr2016lhd}, like the aforementioned WebIsALOD, focuses on the extraction of a hypernym graph. It uses a deep linguistic analysis of the first paragraph in Wikipedia.
    \item \emph{ClaimsKG} \cite{tchechmedjiev2019claimskg} extracts claims from fact checking Web pages,  such as politifact, and interlinks them with other knowledge graphs such as DBpedia, which also allows for finding related claims.
\end{itemize}
The methods discussed in this chapter can be used to assess those emerging knowledge graphs and discuss their added value over existing ones. So, for example, for the above mentioned DBkWik, we have shown that it is highly complimentary to DBpedia: 95\% of all entities in DBkWik are not contained in DBpedia and vice versa.

In summary, knowledge graphs are a useful ingredient to XAI systems, as they provide ready-to-use cross-domain knowledge.
With this chapter, we have given an overview of existing knowledge graphs on the Web, and some guidelines on picking one or more such graphs to build an application for a task at hand.

\bibliographystyle{plain}
\bibliography{11-heist}

\pagebreak
\appendix
\section{Data sources for the comparison of knowledge graphs}
For the comparison of the knowledge graphs, the following data sources have been used:
\subsection{DBpedia} Version 2016-10
\begin{itemize}
\item \url{http://downloads.dbpedia.org/2016-10/dbpedia_2016-10.owl}
\item \url{http://downloads.dbpedia.org/2016-10/core-i18n/en/instance_types_en.ttl.bz2}
\item \url{http://downloads.dbpedia.org/2016-10/core-i18n/en/instance_types_transitive_en.ttl.bz2}
\item \url{http://downloads.dbpedia.org/2016-10/core-i18n/en/interlanguage_links_en.ttl.bz2}
\item \url{http://downloads.dbpedia.org/2016-10/core-i18n/en/labels_en.ttl.bz2}
\item \url{http://downloads.dbpedia.org/2016-10/core-i18n/en/mappingbased_literals_en.ttl.bz2}
\item \url{http://downloads.dbpedia.org/2016-10/core-i18n/en/mappingbased_objects_en.ttl.bz2}
\end{itemize}

\subsection{YAGO} Version 3.1
\begin{itemize}
\item \url{http://resources.mpi-inf.mpg.de/yago-naga/yago3.1/yagoTransitiveType.ttl.7z}
\item \url{http://resources.mpi-inf.mpg.de/yago-naga/yago3.1/yagoSchema.ttl.7z}
\item \url{http://resources.mpi-inf.mpg.de/yago-naga/yago3.1/yagoTypes.ttl.7z}
\item \url{http://resources.mpi-inf.mpg.de/yago-naga/yago3.1/yagoTaxonomy.ttl.7z}
\item \url{http://resources.mpi-inf.mpg.de/yago-naga/yago3.1/yagoLiteralFacts.ttl.7z}
\item \url{http://resources.mpi-inf.mpg.de/yago-naga/yago3.1/yagoLabels.ttl.7z}
\item \url{http://resources.mpi-inf.mpg.de/yago-naga/yago3.1/yagoDateFacts.ttl.7z}
\item \url{http://resources.mpi-inf.mpg.de/yago-naga/yago3.1/yagoFacts.ttl.7z}
\item \url{http://resources.mpi-inf.mpg.de/yago-naga/yago3.1/yagoDBpediaInstances.ttl.7z}
\end{itemize}

\subsection{Wikidata} Version 20190628
\begin{itemize}
    \item \url{https://dumps.wikimedia.org/wikidatawiki/entities/latest-truthy.nt.gz}
\end{itemize}

\subsection{BabelNet} Version 3.6
\\ \\
BabelNet Version 3.6 is not publicly available for download, but has been provided by the developers upon request.

\subsection{NELL} Version 0.3\#1100
\begin{itemize}
    \item \url{http://wdaqua-nell2rdf.univ-st-etienne.fr/archive/NELL2RDF_0.3_1100_ontology.ttl.gz}
    \item \url{http://wdaqua-nell2rdf.univ-st-etienne.fr/archive/NELL2RDF_0.3_1100.nt.gz}
\end{itemize}

\subsection{OpenCyc} Version 4.0
\begin{itemize}
    \item \url{https://github.com/asanchez75/opencyc/blob/master/opencyc-latest.owl.gz}
\end{itemize}

\subsection{CaLiGraph} Version 1.0.6
\begin{itemize}
    \item \url{https://zenodo.org/api/files/042e92ff-0bf7-4d3a-8edb-a5bf491f9a19/caligraph-ontology.nt.bz2}
    \item \url{https://zenodo.org/api/files/042e92ff-0bf7-4d3a-8edb-a5bf491f9a19/caligraph-ontology_dbpedia-mapping.nt.bz2}
    \item \url{https://zenodo.org/api/files/042e92ff-0bf7-4d3a-8edb-a5bf491f9a19/caligraph-ontology_provenance.nt.bz2}
    \item \url{https://zenodo.org/api/files/042e92ff-0bf7-4d3a-8edb-a5bf491f9a19/caligraph-instances_types.nt.bz2}
    \item \url{https://zenodo.org/api/files/042e92ff-0bf7-4d3a-8edb-a5bf491f9a19/caligraph-instances_labels.nt.bz2}
    \item \url{https://zenodo.org/api/files/042e92ff-0bf7-4d3a-8edb-a5bf491f9a19/caligraph-instances_relations.nt.bz2}
    \item \url{https://zenodo.org/api/files/042e92ff-0bf7-4d3a-8edb-a5bf491f9a19/caligraph-instances_dbpedia-mapping.nt.bz2}
    \item \url{https://zenodo.org/api/files/042e92ff-0bf7-4d3a-8edb-a5bf491f9a19/caligraph-instances_provenance.nt.bz2}
\end{itemize}

\end{document}